\documentclass[10pt, twocolumn, twoside, journal]{IEEEtran}
\usepackage{cite,amsmath}
\usepackage{amssymb, graphicx, bm, url}
\usepackage{relsize} % for \smaller to work.
\usepackage{multicol}
\usepackage{multirow}
\usepackage{paralist}

\usepackage{color}  % For highlighting 

\begin{document}
% Title.
% ------
\title{Small-footprint Highway Deep Neural Networks for Speech Recognition}
%
% Single address.
% ---------------
\author{Liang~Lu~\IEEEmembership{Member,~IEEE}, Steve~Renals~\IEEEmembership{Fellow,~IEEE}
\footnote{Copyright (c) 2013 IEEE. Personal use of this material is permitted. However, permission to use this material for any other purposes must be obtained from the IEEE by sending a request to pubs-permissions@ieee.org.}
\thanks{Manuscript received -; revised -}
\thanks{Liang Lu is with Toyota Technological Institute at Chicago, and Steve Renals is with The University of Edinburgh, UK; email: {\smaller \tt llu@ttic.edu, s.renals@ed.ac.uk}}%
\thanks{The research was supported by EPSRC Programme Grant grant EP/I031022/1 \emph{Natural Speech Technology} (NST) and the European Union under H2020 project \emph{SUMMA}, grant agreement 688139.}
}

\markboth{IEEE/ACM Transactions on Audio, Speech and Language Processing, VOL XXX, NO. XXX, 2017}{}
%
% For example:
% ------------
%\address{School\\
%	Department\\
%	Address}
%
% Two addresses (uncomment and modify for two-address case).
% ----------------------------------------------------------
%\twoauthors
%  {A. Author-one, B. Author-two\sthanks{Thanks to XYZ agency for funding.}}
%	{School A-B\\
%	Department A-B\\
%	Address A-B}
%  {C. Author-three, D. Author-four\sthanks{The fourth author performed the work
%	while at ...}}
%	{School C-D\\
%	Department C-D\\
%	Address C-D}
%
%\ninept
%
\maketitle
\begin{abstract}
State-of-the-art speech recognition systems typically employ neural network acoustic models. However, compared to Gaussian mixture models, deep neural network (DNN) based acoustic models often have many more model parameters, making it challenging for them to be deployed on resource-constrained platforms, such as mobile devices. In this paper, we study the application of the recently proposed highway deep neural network (HDNN) for training small-footprint acoustic models. HDNNs are a depth-gated feedforward neural network, which include two types of gate functions to facilitate the information flow through different layers.  Our study demonstrates that HDNNs are more compact than regular DNNs for acoustic modeling, i.e., they can achieve comparable recognition accuracy with many fewer model parameters. Furthermore, HDNNs are more controllable than DNNs: the gate functions of an HDNN can control the behavior of the whole network using a very small number of model parameters. Finally, we show that HDNNs are more adaptable than DNNs. For example, simply updating the gate functions using  adaptation data can result in considerable gains in accuracy. We demonstrate these aspects by experiments using the publicly available AMI corpus, which has around 80 hours of training data.  
%Moreover, we also investigate a knowledge distillation technique to further improve the small-footprint HDNN acoustic models.  
 
\end{abstract}


\section{Introduction}
\label{sec:intro}

\IEEEPARstart{D}{eep} Learning has significantly advanced the state-of-the-art in speech recognition over the past few years~\cite{hinton2012deep, seide2011conversational, saon2016ibm}. Most speech recognisers now employ the neural network and hidden Markov model (NN/HMM) hybrid architecture, first investigated in the early 1990s~\cite{bourlard1994connectionist, renals1994connectionist}. Compared to those models, current neural network acoustic models tend to be larger and deeper, made possible by faster computing such as general-purpose graphic processing units (GPGPUs). Furthermore, more complex neural architectures such  as recurrent neural networks (RNNs) with long short-term memory (LSTM) units and convolutional neural networks (CNNs) have received intensive research, resulting in a range of flexible and powerful neural network architectures that have been applied to a range of tasks in speech, image and natural language processing.  

Despite their success, neural network models have been criticized as lacking structure, being resistant to interpretation, and possessing limited adaptablity. Furthermore accurate neural network acoustic models reported in the research literature have tended to be much larger than conventional Gaussian mixture models, thus making it challenging to deploy them on resource constrained embedded or mobile platforms when  cloud computing solutions are not appropriate (due to the unavailability of an internet connection or for privacy reasons). 
%This motivates the development of smaller acoustic models which can still achieve high recognition accuracy with limited computational resource.  
Recently, there has been considerable work to reduce the size of neural network acoustic models while limiting any reduction in recognition accuracy, such as the use of low-rank matrices~\cite{xue2013restructuring, sainath2013low},  teacher-student training~\cite{li2014learning, ba2014deep, romero15_fitnet}, and structured linear layers~\cite{le2013fastfood, sindhwani2015structured, moczulski2015acdc}. Smaller footprint models may also bring advantages in requiring less training data, and in being potentially more adaptable to changing target domains, environments or speakers, owing to having fewer model parameters.

In this paper, we present a comprehensive study of small-footprint acoustic models using highway deep neural networks (HDNNs), building on our previous studies~\cite{llu2016a, lu2016sequence, lu2016knowledge}. HDNNs are multi-layer networks which have shortcut connections between hidden layers~\cite{srivastava2015training}.  Compared to regular multi-layer networks with skip connections, HDNNs are additionally equipped with two gate functions --  {\it transform} and {\it carry} gates -- which control and facilitate the information flow throughout the whole network.  In particular, the transform gate scales the output of a hidden layer, and the carry gate is used to pass through a layer input directly after element-wise rescaling.  These gate functions are central to training very deep networks~\cite{srivastava2015training} and to speeding up convergence~\cite{llu2016a}.  We show that for speech recognition, recognition accuracy can be retained by increasing the depth of the network, while the number of hidden units in each hidden layer can be significantly reduced. As a result, HDNNs are much thinner and deeper with many fewer model parameters. Besides, in contrast to training regular multi-layer networks of the same depth and width, which typically requires careful pretraining, we demonstrate that HDNNs may be trained using standard stochastic gradient descent without any pretraining~\cite{llu2016a}. To further reduce the number of model parameters, we propose a variant of HDNN architecture by sharing the gate units across all the hidden layers. Furthermore, The authors in~\cite{srivastava2015training} only studied the constrained carry gate setting for HDNNs, while in this work we provide detailed comparisons of different gate functions in the context of speech recognition. 

We also investigate the roles of the two gate functions in HDNNs using both cross-entropy (CE) training and sequence training, and We present a different way to investigate and understand the effect of gate units in neural networks from the point of view of regularization and adaptation. Our key observation is that the gate functions can manipulate the behavior of all the hidden layers, and they are robust to overfitting. For instance, if we do not update the model parameters in the hidden layers and/or the softmax layer during sequence training, and only update the gate functions, then we are able to retain most of the improvement by sequence training.  Moreover, the regularization term in the sequence training objective is not required when only updating the gate functions. Since the size of the gate functions are relatively small, we can achieve a considerable gain by only fine tuning these parameters for unsupervised speaker adaptation, which is a strong advantage of this model.  Finally, we investigate teacher-student training, and its combination with sequence training, as well as speaker adaptation to further improve the accuracy of the small-size HDNN acoustic models. Our teacher-student training experiments also provide more results to understand this technique in the sequence training and adaptation setting. 

Overall, a small-footprint HDNN acoustic model with 5 million model parameters achieved slightly better accuracy compared to a DNN system with 30 million parameters, while the HDNN model with 2 million parameters achieved only slightly lower accuracy compared to that DNN system. Finally, the recognition accuracy of a much smaller HDNN model (less than 0.8 million model parameters) can be significantly improved by teacher-student style training, narrowing the gap between this model and the much larger DNN system. 

\section{Highway Deep Neural Networks}
\label{sec:hdnn}

\subsection{Deep neural networks}
We focus on feed-forward deep neural networks (DNNs) in this study. Although recurrent neural networks with long short-term memory units (LSTM-RNNs) and convolutional neural networks (CNNs) can obtain higher recognition accuracy with fewer model parameters compared to DNNs~\cite{sak2014long, abdel2014convolutional}, they are computationally more expensive for applications on resource constrained platforms. Moreover, their accuracy can be transferred to a DNN by teacher-student training~\cite{chan2015transferring, wong2016sequence, chebotar2016distilling}. 

A multi-layer network with $L$ hidden layers is represented as
\begin{align}
\bm h_1 &= \sigma(\bm x, \theta_1) \\
\bm h_l &= \sigma(\bm h_{l-1}, \theta_l), \quad \text{for} \quad l=2,\ldots, L \\
\label{eq:sm}
\bm y &= g(\bm h_{L}, \theta_c)
\end{align}
where: $\bm x$ is an input feature vector;
%to the network at the time step; 
$\sigma(\bm h_t^{(l-1)}, \theta_l)$ denotes the transformation of the input $\bm h_t^{(l-1)}$ with the parameter $\theta_l$ followed by a nonlinear activation function $\sigma$, e.g., {\tt sigmoid}; $g(\cdot, \theta_c)$ is the output function that is parameterized by $\theta_c$ in the output layer, which usually uses the softmax to obtain the posterior probability of each class given the input feature. To facilitate our discussion later on, we denote $\theta_h$=$\{\theta_1, \cdots, \theta_L\}$ as the set of neural network parameters.
%of model parameters in the neural network feature extractor.

Given target labels, the network is usually trained by gradient descent to minimize a loss function such as cross-entropy.  However, as the number of hidden layers increases, the error surface becomes increasingly non-convex, and it becomes more likely to find a poor local minimum using gradient-based optimization algorithms with random initialization~\cite{erhan2009difficulty}. Furthermore the variance of the back-propagated gradients may become small in the lower layers if the model parameters are not initialized properly~\cite{glorot2010understanding}.

\subsection{Highway networks}
There have been a variety of training algorithms, and model architectures, proposed to enable very deep multi-layer networks including pre-training~\cite{hinton2006reducing, bengio2007greedy}, normalised initialisation~\cite{glorot2010understanding}, deeply-supervised networks~\cite{lee2014deeply}, and batch normalisation~\cite{ioffe2015batch}.  Highway deep neural networks (HDNNs)~\cite{srivastava2015training} were proposed to enable very deep networks to be trained by augmenting the hidden layers with gate functions:
\begin{align}
\label{eq:hw}
{\bm h}_l &= \sigma({\bm h}_{l-1}, \theta_l)\circ T({\bm h}_{l-1}, {\bm W}_T) \nonumber \\ 
& \qquad \quad + {\bm h}_{l-1}\circ C({\bm h}_{l-1}, {\bm W}_c)
\end{align}
where: ${\bm h}_l$ denotes the hidden activations of $l$-th layer; $T(\cdot)$ is the {\it transform gate} that scales the original hidden activations; $C(\cdot)$ is the {\it carry gate}, which scales the input before passing it directly to the next hidden layer; and $\circ$ denotes elementwise multiplication.   The outputs of $T(\cdot)$ and $C(\cdot)$ are constrained to be within $[0, 1]$, and we use a sigmoid function for each, parameterized by $\mathbf{W}_T$ and $\mathbf{W}_c$ respectively. Following our previous work~\cite{llu2016a}, we tie the parameters in the gate functions across all the hidden layers, which can significantly save model parameters. Untying the gate functions did not result in any gain in our preliminary experiments. In this work, we do not use any bias vector in the two gate functions. Since the parameters in $T(\cdot)$ and $C(\cdot)$ are layer-independent, we denote $\theta_g=(\bm W_T, \bm W_c)$, and we will look into the specific roles of these model parameters in sequence training and model adaptation experiments. 

Without the transform gate, i.e. $T(\cdot) = \mathbf{1}$, the highway network is similar to a network with skip connections -- the main difference is that the input is firstly scaled by the carry gate. If the carry gate is set to zero, i.e. $C(\cdot) = \mathbf{0}$, the second term in (\ref{eq:hw}) is dropped,
\begin{align}
\label{eq:trans}
\bm h_l = \sigma(\bm h_{l-1}, \theta_l)\circ T(\bm h_{l-1}, \bm W_T),
\end{align}
resulting in a model that is similar to dropout regularization~\cite{hinton2012improving}, which may be written as
\begin{align}
\bm h_l = \sigma(\bm h_{l-1}, \theta_l)\circ \bm \epsilon, \quad \epsilon_i \sim p(\epsilon_i),
\end{align}
where $p(\epsilon_i)$ is a Bernoulli distribution for the $i$-th element in $\pmb{\epsilon}$ as originally proposed in~\cite{hinton2012improving}; it was shown later that using a continuous distribution with well designed mean and variance works as well or better~\cite{srivastava2014dropout}.  From this perspective, the transform gate may work as a regularizer, but with the key difference that $T(\cdot)$ is a deterministic function, while $\epsilon_i$ is drawn stochastically from a predefined distribution in dropout.  The network in \eqref{eq:trans} is also related to LHUC (Learning Hidden Unit Contribution) adaptation for multilayer acoustic models~\cite{swietojanski2014learning,swietojanski2016lhuc}, which may be represented as
\begin{align}
\bm h_l^{s} = a(\bm r_l^s) \circ \sigma(\bm h_{l-1}^s, \theta_l)
\end{align}
where: $\bm r_l^s$ is a speaker dependent vector for $l$-th hidden layer, and $\bm h_l^s$ is the speaker adapted hidden activations; $s$ is the speaker index; and $a(\cdot)$ is a nonlinear function. The model in \eqref{eq:trans} can be seen as an extension of LHUC in which $\bm r_l^s$ is parameterized as $\bm W_T\bm h_{l-1}$.  We shall investigate the update of $\bm W_T$ for speaker adaptation in the experimental section.   

Although there are more computational steps for each hidden layer compared to regular DNNs due to the gate functions, the training speed will be improved if the size of the weight matrices are smaller.  Furthermore, the matrices can be packed together as
\begin{align} 
\tilde{\bm W}_l = \left[ \bm W_l^\top, \bm W_T^\top, \bm W_c^\top \right]^\top,
\end{align}
where $\bm W_l^\top$ is the weight matrix in the $l$-th layer, and we then compute $\tilde{\bm W}_l\bm h_{l-1}$.
% once for all. 
This approach, applied at the minibatch level, allows more efficient matrix computation when using GPUs.

\subsection{Related models}

Both HDNNs and LSTM-RNNs~\cite{hochreiter1997long} employ gate functions. However, the gates in LSTMs are designed to control the information flow through time and to model along temporal dependencies; for HDNNs, the gates are used to facilitate the information flow through the depth of the model. Combinations of the two architectures have been explored recently: highway LSTMs~\cite{zhang2015highway} employ highway connections to train a stacked LSTM with multiple layers; recurrent highway networks~\cite{Zilly16_recurrent} share gate functions to control the information flow in both time and model depth. On the other hand, the residual network (ResNet)~\cite{he2015deep} was recently proposed to train very deep networks, advancing the state-of-the-art in computer vision. ResNets are closely related to highway networks in the sense that they also rely on skip connections for training very deep networks; however,  gate functions are not employed in ResNets (which can save some computational cost). Finally, adapting approaches developed for visual object recognition \cite{simonyan2014very}, very deep CNN architectures have been investigated for speech recognition~\cite{qian2016very}.

\section{Training}
\label{sec:train}

\subsection{Cross-entropy training}
The most common criterion used to train neural networks for classification is the cross-entropy (CE) loss function,
\begin{align}
\label{eq:ce0}
\mathcal{L}^{(CE)}(\theta) = - \sum_j \hat{y}_{jt} \log y_{jt}, 
\end{align}
where $j$ is the index of the hidden Markov model (HMM) state, $\bm y_t$ is the output of the neural network \eqref{eq:sm} at  time  $t$, and $\hat{\bm y}_t=\{y_{1t}, \cdots, y_{Jt}\}$ denotes the ground truth label that is a one-hot vector, where $J$ is the number of HMM states. Note that the loss function is defined for one training example here for simplicity of notation. Supposing that $ \hat{y}_{jt} = \delta_{ij}$, where $\delta_{ij}$ is the Kronecker delta function and $i$ is the ground truth class at the time step $t$,  the CE loss becomes 
\begin{align}
\label{eq:ce}
\mathcal{L}^{(CE)}(\theta) = - \log y_{it}.
\end{align}
In this case, minimizing $\mathcal{L}^{(CE)}(\theta)$ corresponds to minimizing the negative log posterior probability of the correct class, and  is equal to maximizing the probability $y_{it}$; this will also result in minimizing the posterior probabilities of other classes since they sum to one.

\subsection{Teacher-Student training}
Instead of using the ground truth labels, the teacher-student training approach defines the loss function as
\begin{align}
\label{eq:kd}
\mathcal{L}^{(KL)}(\theta) = - \sum_j \tilde{y}_{jt} \log y_{jt},
\end{align}
where $\tilde{y}_{jt}$ is the output of the teacher model, which works as a pseudo-label. Minimizing this loss function is equivalent to minimizing the Kullback-Leibler (KL) divergence between the posterior probabilities of each class from the teacher and student models~\cite{li2014learning}. Here, $\tilde{y}_{jt}$  is no longer a one-hot vector; instead, the competing classes will have small but nonzero posterior probabilities for each training example.  Hinton et al.~\cite{hinton2015distilling} suggested that the small posterior probabilities are valuable information that encode correlations among different classes. However, their roles may be very small in the loss function as these probabilities are close to zero due to the softmax function. To address this problem, a temperature parameter, $T\in \mathbb{R}^+$, may be used to flatten the posterior distribution,
\begin{align}
y_{jt} &= \frac{\exp \left(z_{jt}/T\right)}{\sum_{i} \exp \left(z_{it}/T\right)}, \\
\bm z_t &= \bm W_{L+1}\bm h_{Lt} + \bm b_{L+1},
\end{align}
where $\bm W_{L+1}, \bm b_{L+1}$ are parameters in the softmax layer.  Following~\cite{hinton2015distilling}, we applied the same temperature to the softmax functions in both the teacher and student networks in our experiments.\footnote{Only increasing the temperature in the teacher network resulted in much higher error rates in pilot experiments.}

A particular advantage of  teacher-student training  is that unlabelled  data can be used easily. However, when  ground truth labels are available, the two loss functions can be interpolated to give a hybrid loss parametrised by $q\in \mathbb{R}^+$ 
\begin{align}
\label{eq:hybrid}
\widetilde{\mathcal{L}(\theta)} = \mathcal{L}^{(KL)}(\theta) + q\mathcal{L}^{(CE)}(\theta) .
\end{align}

\subsection{Sequence training}

While the previous two loss functions are defined at the frame level, sequence training defines the loss at the sequence level, which usually yields a significant improvement in speech recognition accuracy~\cite{kingsbury2012scalable, Vesely:IS13, su2013error}. Given a sequence of acoustic frames, ${\bm X} = \{\bm x_1, \ldots, \bm x_T\}$, of length $T$, and a sequence of labels, $\bm Y$, then the loss function from the state-level minimum Bayesian risk criterion (sMBR)~\cite{gibson2006hypothesis, kingsbury2009lattice} is defined as
\begin{align}
\label{eq:smbr}
\mathcal{L}^{(sMBR)}(\theta) = \frac{\sum_{\mathcal{W} \in \Phi}p({\bm X} \mid \mathcal{W})^k P(\mathcal{W})A(\bm Y, \hat{\bm Y})}{\sum_{\mathcal{W} \in \Phi}p({\bm X} \mid \mathcal{W})^k P(\mathcal{W})},
\end{align}
where: $A(\bm Y, \hat{\bm Y})$ measures the state-level distance between the ground truth and predicted labels; $\Phi$ denotes the hypothesis space represented by a denominator lattice;  $\mathcal{W}$ is the word-level transcription; and $k$ is the acoustic score scaling parameter. In this paper, we  focus on the sMBR criterion for sequence training since it can achieve comparable or slightly better results than training using the maximum mutual information (MMI) or minimum phone error (MPE) criteria~\cite{Vesely:IS13}.

Only applying the sequence training criterion without regularization may lead to overfitting~\cite{Vesely:IS13,su2013error}. To address this problem, we interpolate the sMBR loss function with the CE loss~\cite{su2013error}, with smoothing parameter $p \in \mathbb{R}^{+}$, 
\begin{align}
\label{eq:reg}
\mathcal{L}(\theta) = \mathcal{L}^{(sMBR)}(\theta) + p \mathcal{L}^{(CE)}(\theta).
\end{align}
A motivation for this interpolation is that the acoustic model is usually first trained using CE, and then fine tuned using sMBR for a few iterations. However, in the case of teacher-student training for knowledge distillation, the model is first trained with the KL loss function \eqref{eq:kd}. Hence, we apply the following interpolation when switching from the KL loss function \eqref{eq:kd} to the sequence-level loss function in the case of teacher-student training:
\begin{align}
\label{eq:regts}
\widehat{\mathcal{L}(\theta)} = \mathcal{L}^{(sMBR)}(\theta) + p \mathcal{L}^{(KL)}(\theta).
\end{align}
Again, $p\in\mathbb{R}^+$ is the smoothing parameter, and we have used the same ground truth labels $\bm Y$ when measure the sMBR loss as  in the the standard sequence training .
 
\subsection{Adaptation}
\label{sec:spk}
Adaption of deep neural networks is challenging due to the large number of unstructured model parameters and the small amount of adaptation data. However, the HDNN architecture is more structured as the parameters in the gate functions are layer-independent, and 
%as will be demonstrated further, they are able to 
can control the behavior of all the hidden layers. This motivates the investigation of the adaptation of highway gates by only fine tuning these model parameters.  Although the number of parameters in the gate functions is still  large compared to the amount of per-speaker adaptation data, the size of the gate functions may be controlled by reducing the number of hidden units, but maintaining the accuracy by increasing the depth~\cite{llu2016a}. Moreover, speaker adaptation can be applied to teacher-student training to further improve the accuracy of the compact HDNN acoustic models.

\section{Experiments}
\label{sec:exp}

\subsection{System setup}
Our experiments were performed on the individual headset microphone (IHM) subset of the AMI meeting speech transcription corpus~\cite{renals2007recognition,renals2016ami}.\footnote{\url{http://corpus.amiproject.org}} The amount of training data is around 80 hours, corresponding to roughly 28 million frames. We used 40-dimensional fMLLR adapted features vectors normalised at a per-speaker level, which were then spliced by a context window of 15 frames (i.e. $\pm7$) for all the systems. The number of tied HMM states is 3972, and all the DNN systems were trained using the same alignment. The results reported in this paper were obtained using the CNTK toolkit~\cite{yu2014introduction} with the Kaldi decoder~\cite{povey2011kaldi}, and the networks were trained using the cross-entropy (CE) criterion without pre-training unless specified otherwise. We set the momentum to be 0.9 after the 1st epoch, and we used the sigmoid activation for the hidden layers. The weights in each hidden layer were randomly initialized with a uniform distribution in the range of $[-0.5, 0.5]$ and the bias parameters were initialized to be $0$ for CNTK systems. We used a trigram language model for decoding. 

\begin{table}[t]
\caption{Comparison of DNN and HDNN system with CE and sMBR training. The DNN systems were built using Kaldi, where the networks were pretrained using stacked restricted Boltzmann machines. Results are shown in terms of word error rates (WERs). We use $H$ to denote the size of hidden units, and $L$ the number of layers. $M$ indicates million model parameters.} \vskip 1mm
\label{tab:base}
\centering \small
\begin{tabular}{llc|cc|cc}
\hline 

\hline
  & & & \multicolumn{2}{c|}{ \tt dev} & \multicolumn{2}{c}{\tt eval}  \\
ID & Model  & Size & CE & sMBR & CE & sMBR  \\ \hline
1 & DNN-$H_{2048}L_{6}$ & $30 M$ & 26.0 & 24.3 & 26.8  & 24.6  \\
2 & DNN-$H_{512}L_{10}$ & $4.6 M$ & 26.8 & 25.1 & 28.0 & 25.6  \\
3 & DNN-$H_{256}L_{10}$ & $1.7 M$ & 28.4 & 26.5 & 30.4 & 27.5 \\ 
4 & DNN-$H_{128}L_{10}$ & $0.71 M$ & 31.5 & 29.3 & 34.1 & 30.8  \\ \hline
5 &  HDNN-$H_{512}L_{15}$ & $6.4 M$ & 25.8 & 24.3 & 27.1 & 24.7  \\
6 & HDNN-$H_{512}L_{10}$ & $5.1 M$ & 26.0 & 24.5 & 27.2 & 24.9  \\
7 & HDNN-$H_{256}L_{15}$ & $2.1 M$ & 26.9 & 25.2 & 28.4 & 25.9  \\ 
8 & HDNN-$H_{256}L_{10}$ & $1.8 M$ & 27.2 & 25.2 & 28.6 & 26.0  \\
9 &HDNN-$H_{128}L_{10}$ & $0.74 M$  & 29.9 & 28.1 & 32.0 & 29.4 \\ \hline

 \hline
\end{tabular}
\vskip-4mm
\end{table}

\subsection{Baseline results}

Table \ref{tab:base} shows the CE and sequence training results for baseline DNN and HDNN models of various size. The DNN systems were all trained using Kaldi with RBM pretraining (without pretraining, training thin and deep DNN models did not converge using CNTK). However, we were able to train HDNNs with random initialization without pretraining, demonstrating that the gate functions in HDNNs facilitate the information flow through the layers. For sequence training, we performed the sMBR update for 4 iterations, and set $p=0.2$ in Eq. \eqref{eq:reg} to avoid overfitting. Table \ref{tab:base} shows that the HDNNs achieved consistently lower WERs compared to the DNNs; the margin of the gain also increases as the number of hidden units becomes smaller.  As the number of hidden units decreases, the accuracy of DNNs degrades rapidly, and the accuracy loss cannot be compensated by increasing the depth of the network. The results also show that sequence training improves the recognition accuracy comparably for both DNN and HDNN systems, and the improvements are consistent for both {\tt dev} and {\tt eval} sets. Overall, the HDNN model with around 6 million model parameters has a similar accuracy to the regular DNN system with 30 million model parameters. 

\subsection{Transform and Carry gates}
\label{sec:gate}

\begin{figure}[t]
\small
\centerline{\includegraphics[width=0.5\textwidth]{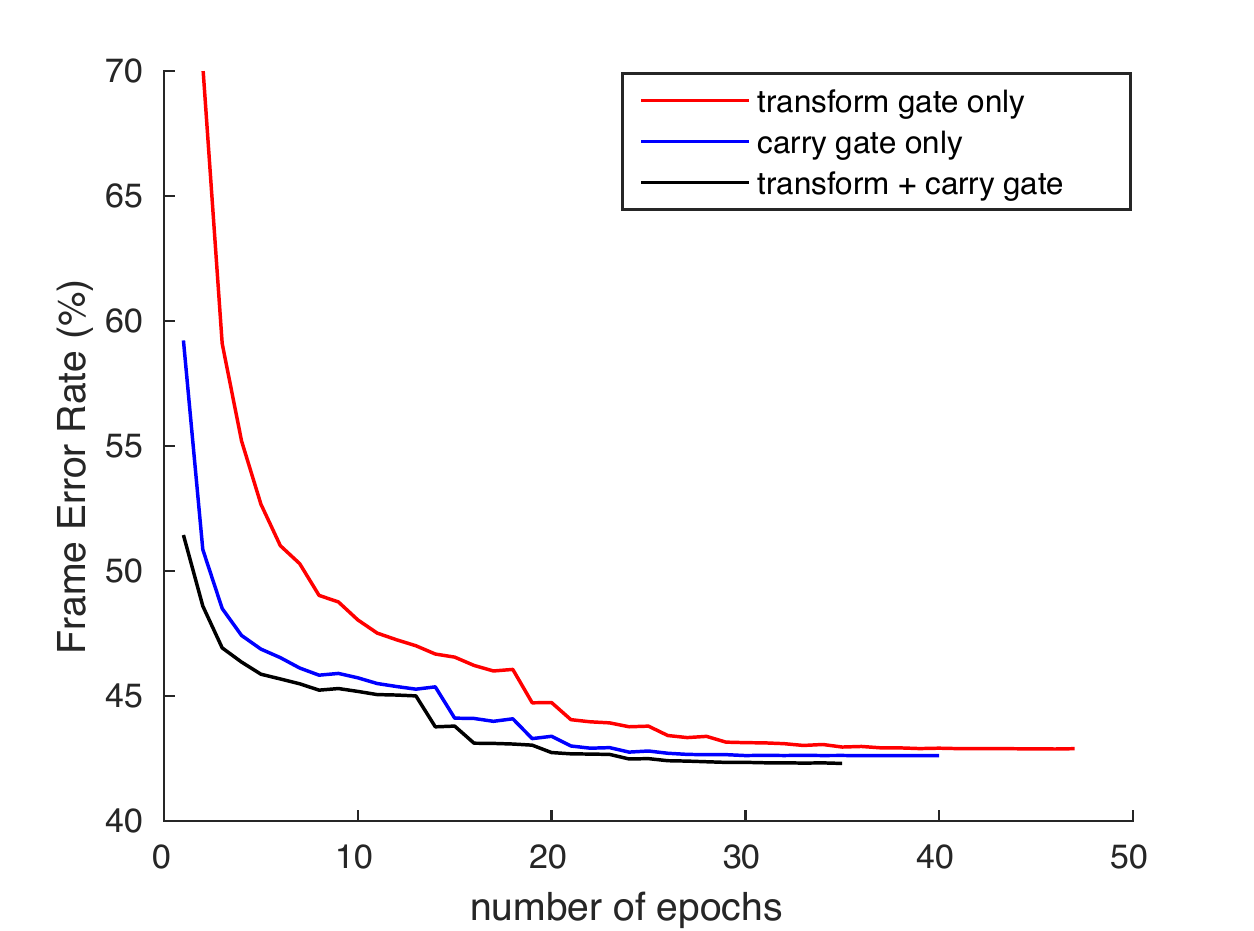}} \vskip -2mm
\caption{Convergence curves for training HDNNs with and without the transform and the carry gate. The frame error rates (FERs) were obtained using the validation dataset.}  
\label{fig:gate}
\end{figure}

We then evaluated the specific role of the transform and carry gates in the highway architectures. The results are shown in Table \ref{tab:gate}, where we disabled each of the gates in turn. We can see that using only one of the two gates, the HDNN can still achieve lower WER compared to the regular DNN baseline, but the best results are obtained when both gates are active, indicating that the two gating functions are complementary. Figure \ref{fig:gate} shows the convergence curves of training HDNNs with and without the transform and carry gates. We observe faster convergence when both gates are active, with considerably slower convergence when using only the transform gate.  This indicates that the carry gate, which controls the skip connections, is more important to the convergence rate. We also investigated constrained gates, in which $C(\cdot) = \mathbf{1} - T(\cdot)$~\cite{srivastava2015training}, which reduces the computational cost since the matrix-vector multiplication for the carry gate is not required.  We evaluated this configuration with 10-layer neural networks, and the results are also shown in Table \ref{tab:gate}: this approach does not improve recognition accuracy in our experiments. 

 \begin{table}
\caption{Results of highway networks with and without the transform and the carry gate. The HDNN-$H_{512}L_{10}$ with both  gates active corresponds to the CE baseline in Table \ref{tab:base}}\vskip 1mm
\label{tab:gate}
\centering \small
\begin{tabular}{l|cccc}
\hline 

\hline
Model & Transform & Carry & Constrained & WER  \\ \hline
%DNN$^*$    & 10 & 512 & $\times$ & $\times$ & 28.0 \\
 & $\surd$ & $\surd$ & $\times$ &  27.2 \\
HDNN-$H_{512}L_{10}$ & $\surd$ & $\times$ & $\times$ &  27.6 \\
& $\times$ & $\surd$ & $\times$ & 27.5 \\
& $\surd$ & $\surd$ & $\surd$ & 27.4 \\

\hline

\hline
\end{tabular}
\vskip-3mm
\end{table}

\begin{table}[t]
\caption{Results of updating of specific sets of model parameters in sequence training (after CE training). $\theta_h$ denotes the hidden layer weights, $\theta_g$ denotes the gating parameters, and $\theta_c$ denotes the parameters in the output softmax layer. CE regularization was used in these experiments.} \vskip 1mm
\label{tab:seq2}
\centering \small
\begin{tabular}{l|ccc|c}
\hline 

\hline
& \multicolumn{3}{c|}{sMBR Update}  &  WER \\
Model    & $\theta_h$ & $\theta_g$ & $\theta_c$  & (\tt eval) \\ \hline
 & $\times$ & $\times$ & $\times$ & 27.2 \\ 
HDNN-$H_{512}L_{10}$ & $\surd$ & $\surd$ & $\surd$ & 24.9 \\ 
  &  $\times$ & $\surd$ & $\surd$ & 25.2\\ 
 & $\times$ & $\surd$ & $\times$ & 25.8\\ \hline
  & $\times$ & $\times$ & $\times$ & 28.6 \\ 
HDNN-$H_{256}L_{10}$  &  $\surd$ & $\surd$ & $\surd$ & 26.0\\ 
  & $\times$ & $\surd$ & $\surd$ & 26.6 \\ 
 & $\times$ & $\surd$ & $\times$ & 27.0 \\ \hline
  & $\times$ & $\times$ & $\times$ & 27.1 \\ 
HDNN-$H_{512}L_{15}$  &  $\surd$ & $\surd$ & $\surd$ & 24.7 \\ 
  & $\times$ & $\surd$ & $\surd$ & 25.2 \\ 
 & $\times$ & $\surd$ & $\times$ & 25.6 \\ \hline
   & $\times$ & $\times$ & $\times$ & 28.4 \\ 
HDNN-$H_{256}L_{15}$  &  $\surd$ & $\surd$ & $\surd$ & 25.9 \\ 
  & $\times$ & $\surd$ & $\surd$ & 26.4 \\ 
 & $\times$ & $\surd$ & $\times$ & 26.6 \\ \hline
  
 \hline
\end{tabular}
\end{table}

To look into the relative importance of the gate functions to other type of model parameters in the feature extractor and classification layer,  we also performed a set of ablation experiments with sequence training, where we removed the update of different sets of model parameters (after CE training). These results are given in Table \ref{tab:seq2}, which shows that only updating the parameters in the gates $\theta_g$ can retain most of the improvement given by sequence training, while updating $\theta_g$ and $\theta_c$ can achieve the accuracies close to the optimum. Although $\theta_g$ only accounts for a small fraction of the total number of parameters (e.g., $\sim$ 10\% for the HDNN-$H_{512}L_{10}$ system and $\sim$ 7\% for the HDNN-$H_{256}L_{10}$ system), the results demonstrate that it plays an important role in manipulating the behavior of the neural network feature extractor. 

\begin{table}[t]
\caption{Results of sMBR training with and without regularization. } \vskip 1mm
\label{tab:smbr2}
\centering \small
\begin{tabular}{l|c|ccc}
\hline 

\hline
& &  \multicolumn{3}{c}{WER ({\tt eval})}   \\
Model   & sMBR Update  & CE  & $p=0.2$ & $p=0$   \\ \hline
HDNN-$H_{512}L_{10}$& $\{\theta_h, \theta_g, \theta_c \}$  & 27.2 & 24.9 & 25.0  \\
HDNN-$H_{512}L_{10}$&  $\theta_g$ & 27.2 & 25.8 & 25.3  \\
HDNN-$H_{256}L_{10}$ & $\{\theta_h, \theta_g, \theta_c \}$  & 28.6 & 26.0 &  28.3 \\ 
HDNN-$H_{256}L_{10}$&  $\theta_g$ & 28.6 & 27.0 & 26.8  \\ \hline

 \hline
\end{tabular}
\end{table}

\begin{figure*}
\small
\centerline{\includegraphics[width=0.85\textwidth]{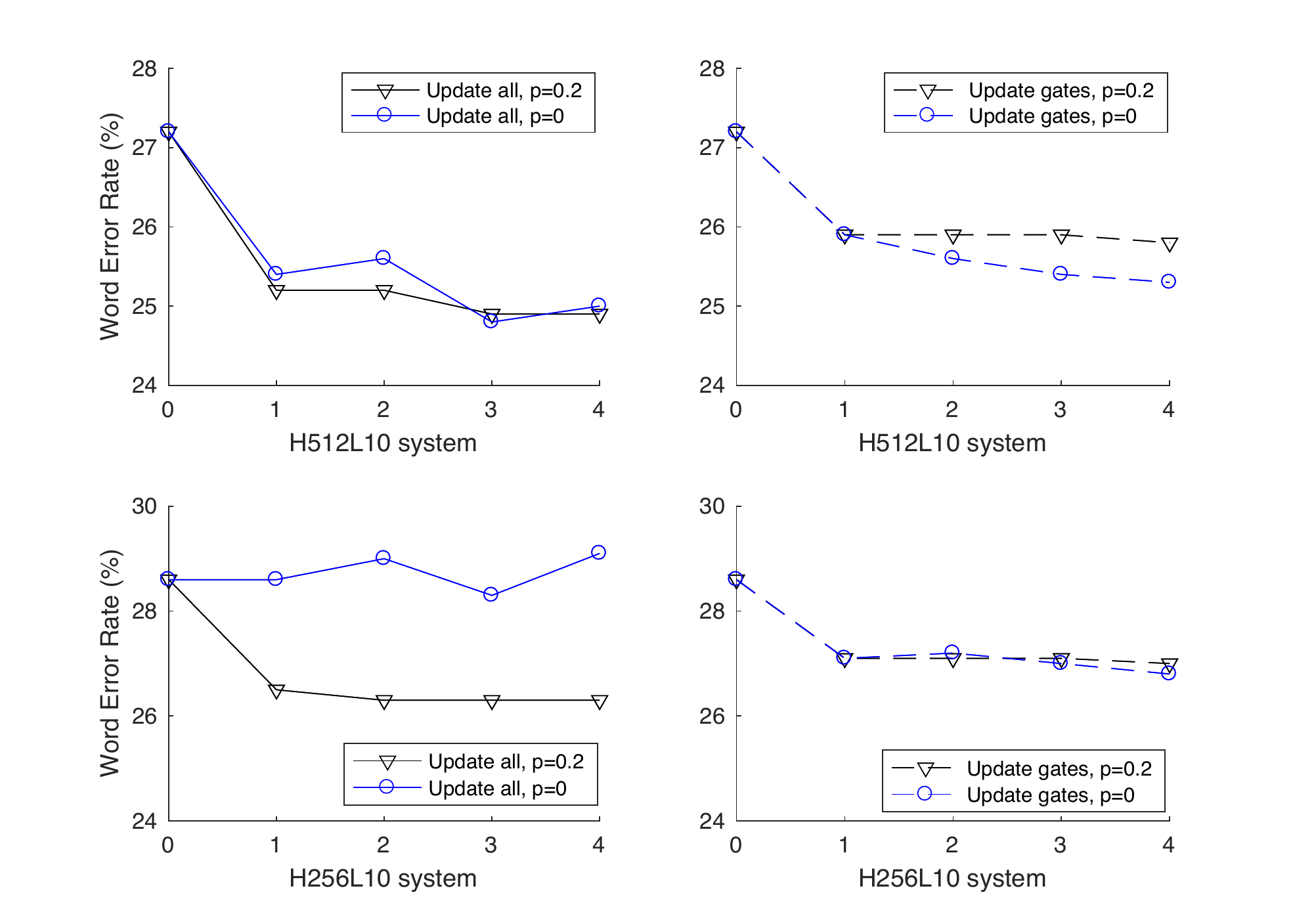}} 
\caption{Convergence curves of sMBR training with and without CE regularization (controlled by parameter $p$). Networks had 10 hidden layers and 256 or 512 hidden units per layer.}
%The regularisation term can stabilise the convergence when updating all the model parameter, while its role is diminishing when updating the gate functions only.} 
\vskip-2mm
\label{fig:reg}
\vskip-2mm
\end{figure*}

Complementary to the above experiments, we then investigated the effect of the regularization term for sequence training of HDNNs \eqref{eq:reg}. We performed the experiments with and without the CE regularization for two system settings, i.e.: i) update all the model parameters; ii) update only the gate functions. Our motivation was to validate if only updating the gate parameters is more resistant to overfitting. The results are given in Table \ref{tab:smbr2}, from which we see that by removing the CE regularization term, we achieved slightly lower WER when updating the gate functions only. However, when updating all model parameters, the regularization term was an important stabilizer for the convergence. Figure \ref{fig:reg} shows the convergence curves for the two system settings. Overall, although the gate functions can largely control the behavior of the highway networks, they are not prone to overfitting when other model parameters are switched off. 

\subsection{Adaptation}
\label{sec:exp-adp}
The previous experiments show that the gate functions can largely control the behavior of a multi-layer neural network feature extractor with a relatively small number of model parameters. This observation inspired us to study speaker adaptation using the gate functions. Our first experiments explored unsupervised speaker adaptation, in which we decoded the evaluation set using the speaker-independent models, and then used the resulting pseudo-labels to fine-tune the gating parameters ($\theta_g$) in the second pass.  The evaluation set contained around 8.6 hours of audio, with 63 speakers, an average of around 8 minutes of speech per speaker, which corresponds to about 50\,000 frames. This is a relatively small amount of adaptation data, given the size of $\theta_g$ (0.5 million parameters in the HDNN-$H_{512}L_{10}$ system). We set the learning rate to be $2\times 10^{-4}$ per sample, and we updated $\theta_g$ for 5 adaptation epochs. 

\begin{figure}[h]
\small
\centerline{\includegraphics[width=0.5\textwidth]{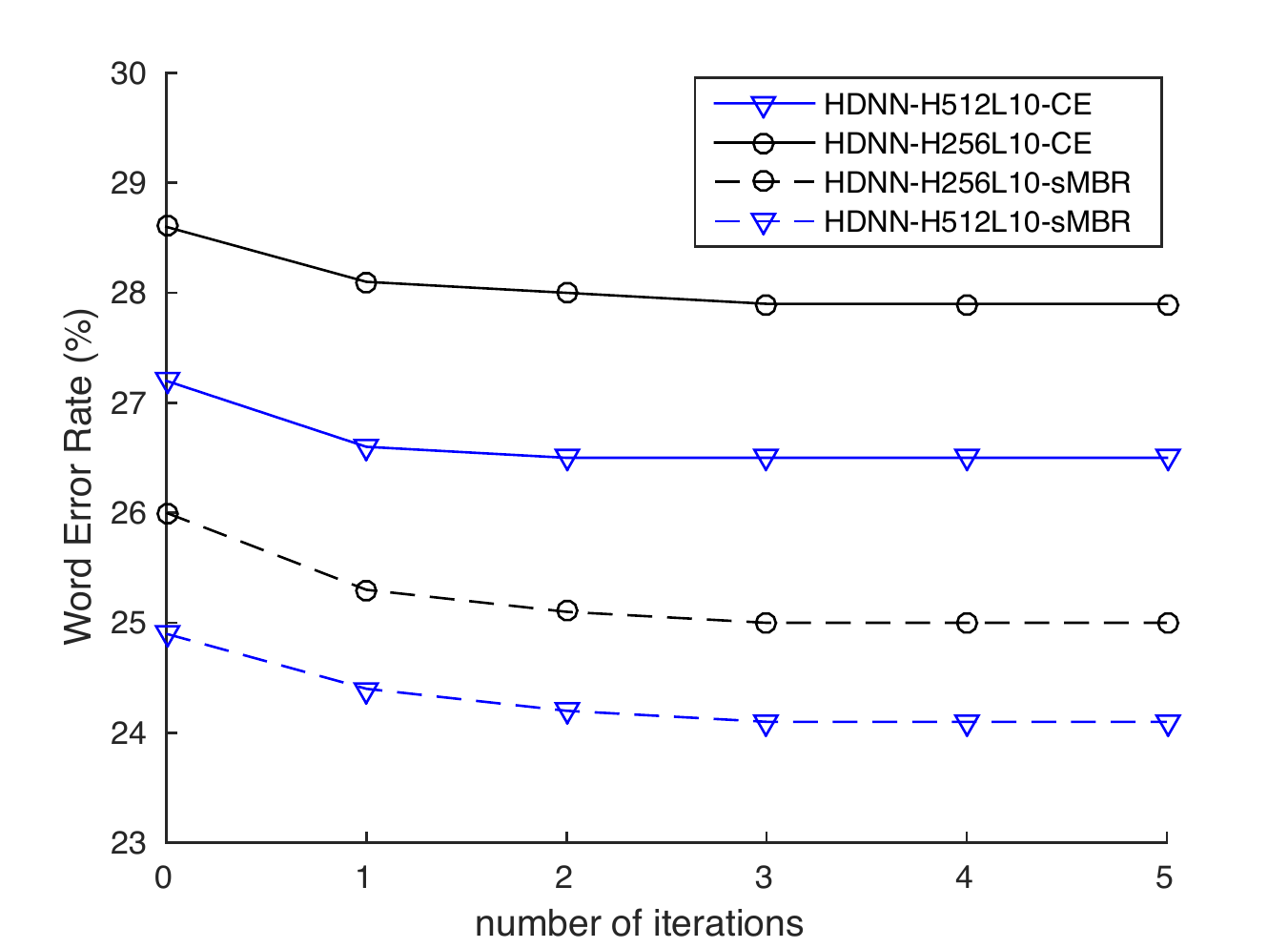}} 
\caption{Unsupervised adaptation results with different number of iterations. The speaker-independent models were trained by CE or sMBR, and we used the CE criterion for all adaptation experiments.}  
\label{fig:adapt}
\end{figure}

Table \ref{tab:adapt} shows the adaptation results, from which we observe a small but consistent reduction in WER  for different model configurations (both CE and sMBR trained) when using fMLLR speaker adapted features.  The results indicate that updating all the model parameters yields smaller improvements. With speaker adaptation and sequence training, the HDNN system with 5 million model parameters (HDNN-$H_{512}L_{10}$) works slightly better than the DNN baseline with 30 million parameters (24.1\%  from row 5 of Table \ref{tab:adapt} vs. 24.6\% from row 1 of Table~\ref{tab:base}), while the HDNN model with 2 million parameters (HDNN-$H_{256}L_{10}$) has only a slightly higher WER compared to the baseline (25.0\% from row 6 of Table \ref{tab:adapt}  vs. 24.6\% from row 1 of Table~\ref{tab:base}). In Figure \ref{fig:adapt} we show the adaptation results for a different number of iterations. We observe that the best results can be achieved after 2 or 3 adaptation iterations; further updating the gate functions $\theta_g$ does not result in overfitting. For validation we  performed experiments with 10 adaptation iterations, and again we did not observe overfitting.  This observation is in line with the sequence training experiments, demonstrating that the gate functions are relatively resistant to overfitting.

\begin{table}[t]
\caption{Results of unsupervised speaker adaptation. Here, we only updated $\theta_g$ using the CE criterion, while the speaker-independent (SI) models were trained by either CE or sMBR. SA denotes speaker adapted models. } \vskip 1mm
\label{tab:adapt}
\centering \small
\begin{tabular}{ll|c|c|cc}
\hline 

\hline
& & & & \multicolumn{2}{c}{WER ({\tt eval})}   \\
ID & Model   & Seed   & Update & SI & SA   \\ \hline
1& HDNN-$H_{512}L_{10}$&   & & 27.2 & 26.5  \\
2& HDNN-$H_{256}L_{10}$ & CE  & & 28.6 & 27.9 \\
3& HDNN-$H_{512}L_{15}$ &   & & 27.1 & 26.4  \\ 
4& HDNN-$H_{256}L_{15}$ &   & $\theta_g$ & 28.4 &  27.6 \\ \cline{1-3}\cline{5-6}

5 & HDNN-$H_{512}L_{10}$ &   & & 24.9 & {\bf 24.1} \\
6& HDNN-$H_{256}L_{10}$ &  & & 26.0 & {\bf 25.0}  \\
7 & HDNN-$H_{512}L_{15}$ &  sMBR & & 24.7 &   24.0 \\ 
8 & HDNN-$H_{256}L_{15}$ &   &  & 25.9 & 24.9 \\ 
9 & HDNN-$H_{128}L_{10}$ &   &  & 29.4 & 28.7 \\ \cline{4-6}
10 & HDNN-$H_{512}L_{10}$ &   &  & 24.9 & 24.5 \\
11 & HDNN-$H_{256}L_{10}$ &  & $\{\theta_h, \theta_g, \theta_c\}$& 26.0 & 25.4  \\ 
12 & HDNN-$H_{128}L_{10}$ &  & & 29.4 & 28.8  \\ \hline

 \hline
\end{tabular}
\vskip-4mm
\end{table}

\begin{figure}[h]
\small
\centerline{\includegraphics[width=0.5\textwidth]{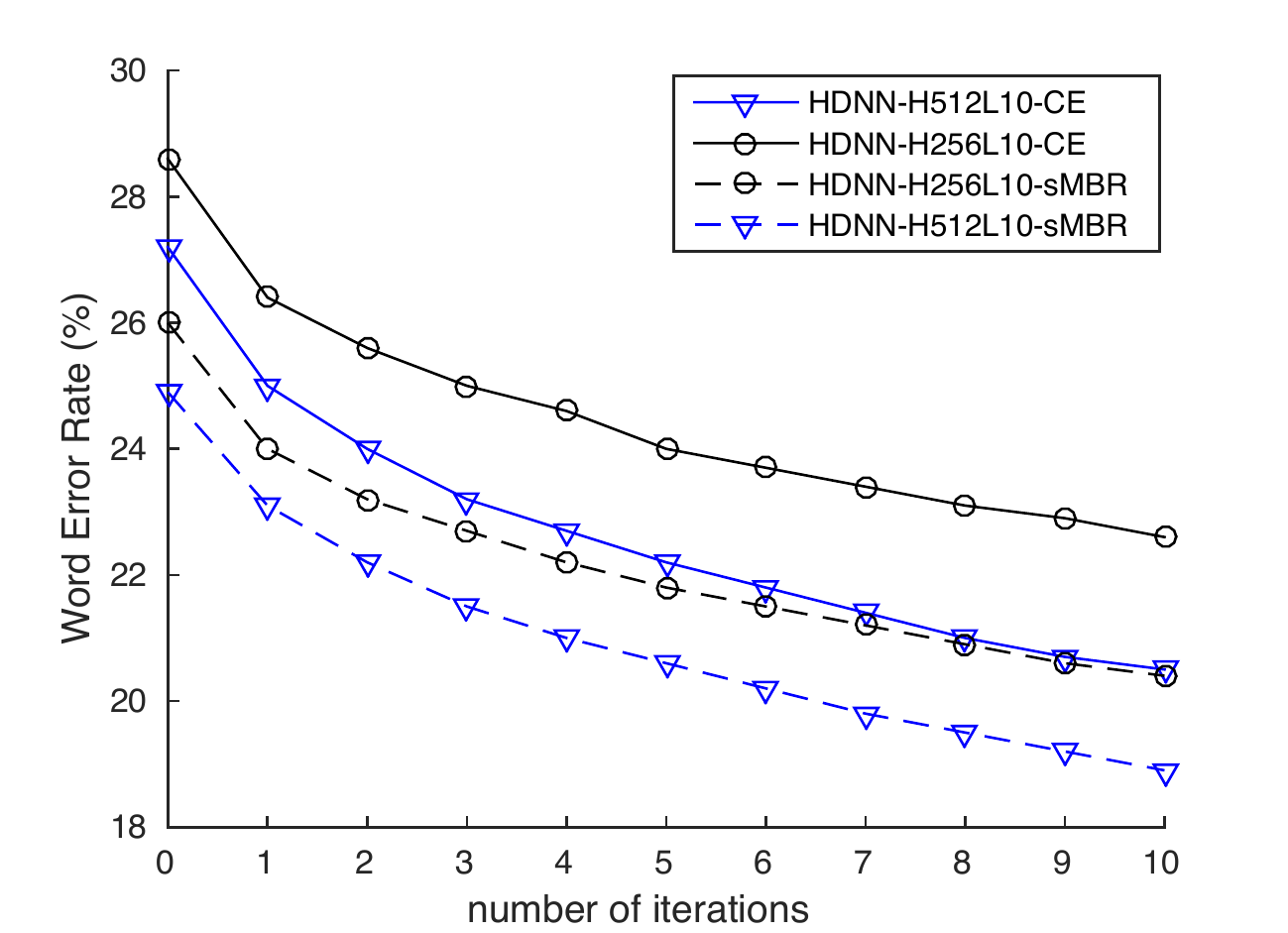}} 
\caption{Supervised adaptation results with oracle labels.}  
\label{fig:oracle}
%\vskip-1mm
\end{figure}

In order to evaluate the impact of the accuracy of the labels to this adaptation method as well as the memorization capacity of the highway gate units, we performed a set of diagnostic experiments, in which we used the oracle labels for adaptation. We obtained the oracle labels from a forced alignment using the DNN model trained with the CE criterion and word level transcriptions. We used this fixed alignment for all the adaptation experiments in order to compare the different seed models. Figure \ref{fig:oracle} shows the adaptation results with oracle labels, suggesting that an increased reduction in WER may be achieved when the supervision labels are more accurate. In the future, we shall investigate the model for domain adaptation, where the amount of adaptation data is usually relatively larger, and the ground truth labels are available. 
%Therefore, the gate functions may have large capacity for adaptation with high quality pseudo labels. To further study this aspect, in the future, we shall investigate supervised adaptation of highway networks.  

\subsection{Teacher-Student training}

\begin{figure}[t]
\small
\centerline{\includegraphics[width=0.5\textwidth]{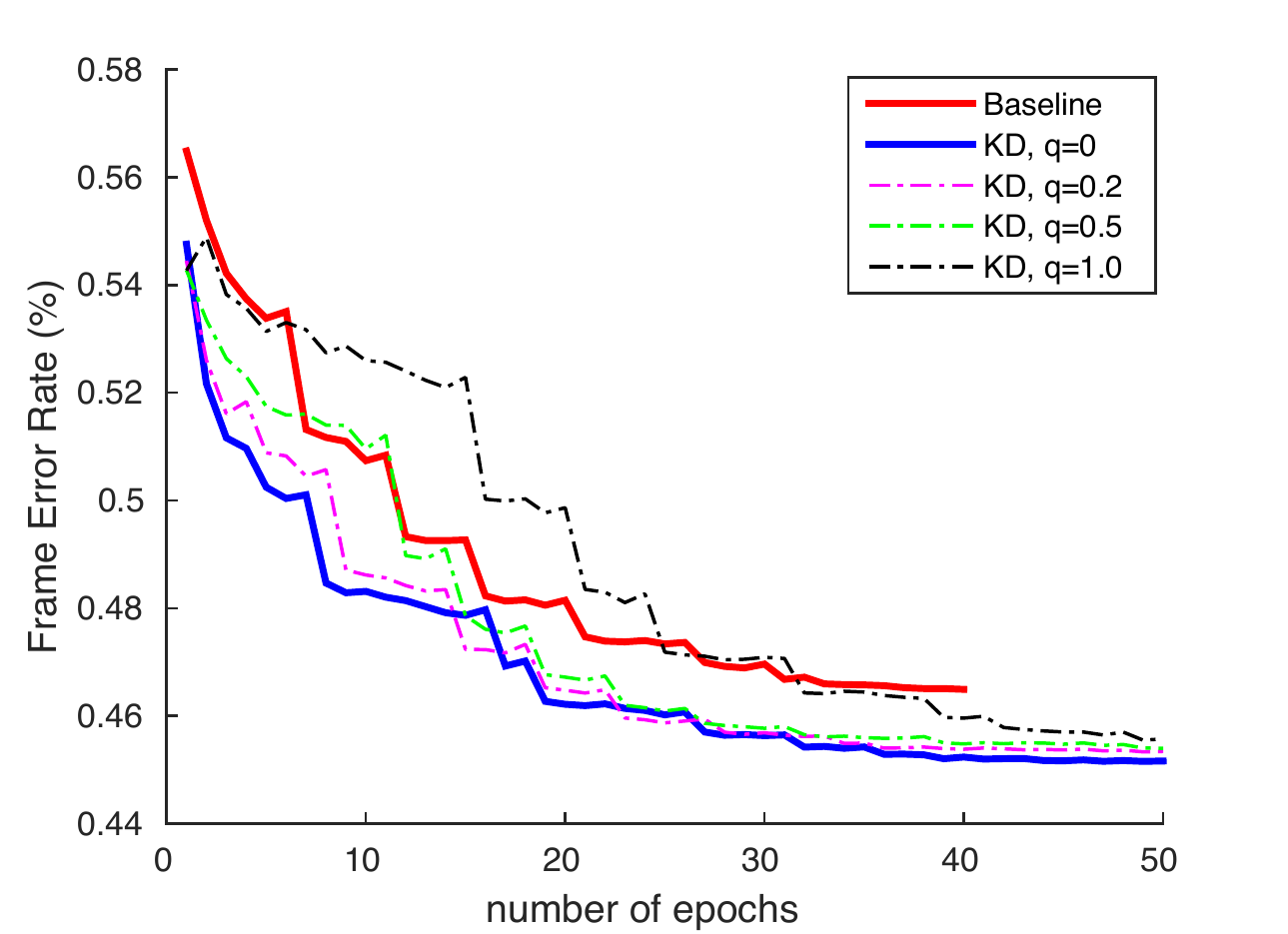}} 
\caption{Convergence curves of teacher-student training. The frame error rates were obtained from the cross validation set.  The convergence slows as $q$ increases. {\tt KD} denotes teacher-student training.}  \vskip-2mm
\label{fig:kd}
%\vskip-2mm
\end{figure}

After sequence training and adaptation, the HDNN with 2 million model parameters has a similar accuracy to the DNN baseline with 30 million model parameters. However, the model HDNN-$H_{128}L_{10}$ which has fewer than 0.8 million model parameters has a substantially higher WER compared to the DNN baseline (28.7\% from row 9 of Table \ref{tab:adapt}  vs. 24.6\% from row 1 of Table~\ref{tab:base}). We  investigated if the accuracy of the small HDNN model can be further improved using teacher-student training. We first compare the teacher-student loss function \eqref{eq:kd} and the hybrid loss function \eqref{eq:hybrid}. We used a CE trained DNN-$H_{2048}L_6$ as the teacher model, and used the HDNN-$H_{128}L_{10}$ as the student model. Figure \ref{fig:kd} shows the convergence curves when training the model with the different loss functions, while Table \ref{tab:kd} shows the WERs. We observe that teacher-student training without the ground truth labels can achieve a significantly lower frame error rate on the cross validation set (Figure \ref{fig:kd}) which corresponds to a moderate WER reduction (Table \ref{tab:kd}: 31.3\% vs. 32.0\% on the {\tt eval} set). However, using the hybrid loss function \eqref{eq:hybrid} does not result in further improvement, and when  $q>0$ during training  convergence is slower (Figure \ref{fig:kd}). We interpret this result as indicating that the probabilities of uncorrected classes may play a lesser role, which supports the argument that they encode useful information for training the student model~\cite{hinton2015distilling}. This hypothesis encouraged us to investigate the use of a high temperature to flatten the posterior probability distribution of the labels from the teacher model. The results are shown in Table \ref{tab:kd}; contrary to our expectation, using high temperatures results in higher WERs. In the following experiments, we fixed $q=0$ and $T=1$. 

\begin{table}[t]
\caption{Results of teacher-student training with different loss functions and temperatures. $q$ denotes the interpolation parameter in Eq. \eqref{eq:hybrid}, and $T$ is the temperature. The teacher models were trained using the CE criterion.} 
\label{tab:kd}
\centering \small
\begin{tabular}{lcc|cc}
\hline 

\hline
  & & & \multicolumn{2}{c}{WER}   \\
Model  & $q$ & $T$ & {\tt eval} & {\tt dev}   \\ \hline
DNN-$H_{128}L_{10}$ & -- & -- & 34.1  & 31.5  \\ 
HDNN-$H_{128}L_{10}$ baseline & -- & -- & 32.0 & 29.9  \\ \hline
HDNN-$H_{128}L_{10}$  & 0 & 1 &  31.3 & 29.3  \\
HDNN-$H_{128}L_{10}$  & 0.2 & 1 & 31.4 & 29.5  \\
HDNN-$H_{128}L_{10}$  & 0.5 & 1& 31.3 & 29.4  \\
HDNN-$H_{128}L_{10}$  & 1.0 & 1 & 31.3 & 29.4 \\ \hline
HDNN-$H_{128}L_{10}$  & 0 & 2 & 32.3 & 29.9 \\ 
HDNN-$H_{128}L_{10}$  & 0 & 3 & 33.0 & 30.6 \\ \hline

 \hline
\end{tabular}
\vskip-2mm
\end{table}

\begin{figure}[t]
\small
\centerline{\includegraphics[width=0.47\textwidth]{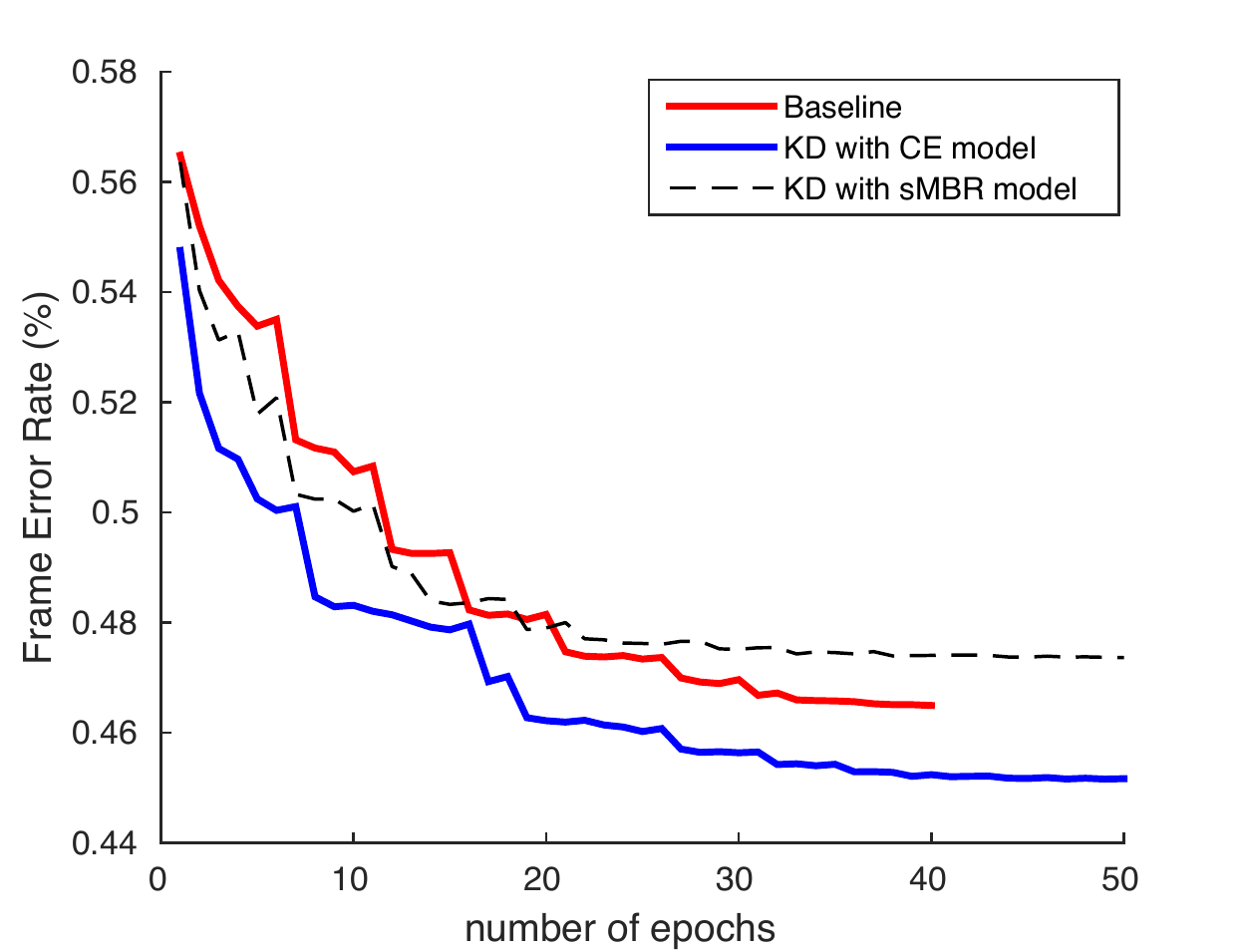}} 
\caption{Convergence curves of teacher-student training with CE or sMBR-based teacher model. }  \vskip-2mm
\label{fig:seq}
%\vskip-2mm
\end{figure}

We then improved the teacher model by sMBR sequence training, and used this model to supervise the training of the student model. We found that the sMBR-based teacher model can significantly improve the performance of the student model (similar to the results reported in~\cite{li2014learning}). In fact, the error rate is lower than that achieved by the student model trained independently with sMBR (28.8\% from row 2 of Table \ref{tab:seq} vs. 29.4\% from row 9 of Table \ref{tab:base} on the {\tt eval} set). Note that, since the sequence training criterion does not maximize the frame accuracy, training the model with this criterion often reduces the frame accuracy (see Figure 6 of~\cite{heigold2014asynchronous}). Interestingly, we observed the same pattern in the case of teacher-student training of HDNNs. Figure \ref{fig:seq} shows the convergence curves of using CE and sMBR based teacher models, where we see that the student model achieves much higher frame error rate on the cross validation set when supervised by sMBR-based teacher model, although the loss function \eqref{eq:kd} is at the frame level.

\begin{table}[t]
\caption{Results of sequence training on the {\tt eval} set. LR denotes the learning rate. The student model is HDNN-$H_{128}L_{10}$. }
\label{tab:seq}
\centering \small
\begin{tabular}{ll|cccc}
\hline 

\hline
ID & Teacher model & LR & $p$ & $\mathcal{L}^{(KL)} \rightarrow \widehat{\mathcal{L}(\theta})$   \\ \hline
1& DNN-$H_{2048}L_{6}$-CE  & $1\times 10^{-5}$ & 0.2 & 31.3 $\rightarrow$ 28.4 \\ % & 29.3$\rightarrow$27.2 \\
2& DNN-$H_{2048}L_{6}$-sMBR& $1\times10^{-5}$ & 0.2 & 28.8 $\rightarrow$ 28.9 \\ %& 27.6$\rightarrow$27.9  \\ 
3 & DNN-$H_{2048}L_{6}$-sMBR& $1\times10^{-5}$ & 0.5 & 28.8 $\rightarrow$ 28.0 \\  % & 27.6$\rightarrow$27.9  \\ \hline
4 & DNN-$H_{2048}L_{6}$-sMBR& $5\times10^{-6}$ & 0.2 & 28.8 $\rightarrow$ 28.6 \\  % & 27.6$\rightarrow$27.9  \\ \hline
5 & DNN-$H_{2048}L_{6}$-sMBR& $5\times10^{-6}$ & 0.5 & 28.8 $\rightarrow$ 28.0 \\ \hline % & 27.6$\rightarrow$27.9  \\ \hline

 \hline
\end{tabular}
\vskip-2mm
\end{table}

We then investigated whether the accuracy of the student model can be further improved by the sequence level criterion. Here, we set the smoothing parameter $p=0.2$ in \eqref{eq:regts} and the default learning rate to be $10^{-5}$ following our previous work~\cite{lu2016sequence}. Table \ref{tab:seq} shows sequence training results for student models supervised by both CE and sMBR-based teacher models. Surprisingly, the student model supervised by the CE-based DNN model can be significantly improved by sequence training -- the WER obtained by this approach is lower compared to the model trained independently with sMBR (28.4\% from row 1 of Table \ref{tab:seq}  vs. 29.4\% from row 9 of Table \ref{tab:base} on the {\tt eval} set). However, this configuration did not improve the student model supervised by an sMBR-based teacher model. After inspection, we found that this was due to overfitting.  We then increased the value of $p$ to enable stronger regularization and reduced the learning rate. Lower WERs were obtained as the table shows; however, the improvement is less significant as the sequence level information has already been integrated into the teacher model.

\begin{table}[t]
\caption{Results of unsupervised speaker adaptation. The hard labels are ground truth labels, and the soft labels are provided by the teacher model. HDNN-$H_{128}L_{10}$-KL denotes the student model.}
\label{tab:adap}
\centering \small
\begin{tabular}{l|cccc}
\hline 

\hline
 &   & & \multicolumn{2}{c}{{\tt eval}}   \\
Model & Label & Update & SI & SA  \\ \hline
HDNN-$H_{128}L_{10}$ & Hard & $\{\theta_h, \theta_g, \theta_c\}$ & 29.4 & 28.8 \\
HDNN-$H_{128}L_{10}$ & Hard & $\theta_g$  & 29.4 & 28.7  \\
HDNN-$H_{128}L_{10}$-KL & Soft & $\{\theta_h, \theta_g, \theta_c\}$   &  28.4 & 27.5 \\ 
HDNN-$H_{128}L_{10}$-KL & Soft & $\theta_g$ & 28.4 & 27.8 \\ 
HDNN-$H_{128}L_{10}$-KL &Hard & $\{\theta_h, \theta_g, \theta_c\}$   & 28.4 & 27.7 \\ 
HDNN-$H_{128}L_{10}$-KL &Hard & $\theta_g$  & 28.4 & 27.1 \\ \hline

 \hline
\end{tabular}
\end{table}

\subsection{Teacher-Student training with adaptation}

We then performed similar adaptation experiments to section \ref{sec:exp-adp} for HDNNs trained by the teacher-student approach. We applied the second-pass adaptation approach for the standalone HDNN model, i.e., we decoded the evaluation utterances to obtain the hard labels first, and then used these labels to adapt the model using the CE loss \eqref{eq:ce}. However, when using the teacher-student loss \eqref{eq:kd} only one-pass decoding is required because the pseudo-labels for adaptation are provided by the teacher, which does not need a word level transcription. This is a particular advantage of the teacher-student training technique. However, for resource-constrained application scenarios, the student model should be adapted offline, because otherwise the teacher model needs to be accessed to generate the labels. This requires another set of unlabelled speaker-dependent data for adaptation, which is usually not expensive to collect.

\begin{table}[h]
\caption{Summary of our results. } 
\label{tab:summary}
\centering \small
\begin{tabular}{l|cccc}
\hline 

\hline

Model & Size & WER   \\ \hline
DNN-$H_{2048}L_{6}$ CE baseline & $30M$ & 26.8 \\ 
$\qquad \qquad$ +sMBR training & $30M$ & {\bf 24.6} \\ \hline
HDNN-$H_{512}L_{10}$ CE baseline & $5.1M$ & 27.2 \\
$ \qquad$ +sMBR training & $5.1M$ & 24.9 \\ 
$\qquad \qquad$ + adaptation & $5.1M$ & {\bf 24.0} \\ \hline
HDNN-$H_{256}L_{10}$ CE baseline & $1.8M$ & 28.6 \\
$ \qquad$ +sMBR training & $1.8M$ & 26.0 \\ 
$\qquad \qquad$ + adaptation & $1.8M$ & {\bf 25.0} \\ \hline
HDNN-$H_{128}L_{10}$ CE baseline & $0.74M$ & 32.0 \\
$ \quad$ +sMBR training & $0.74M$ & 29.4  \\ 
$\qquad$ + teacher-student training & $0.74 M$ & 28.4 \\ 
$\quad \qquad$ + adaptation & $0.74 M$ & {\bf 27.1} \\ \hline

 \hline
\end{tabular}
\vskip-3mm
\end{table}

Since the standard AMI corpus does not have an additional set of speaker-dependent data, we only show online adaptation results. We used the teacher-student trained model from row 1 of Table \ref{tab:seq} as the speaker-independent (SI) model because its pipeline is much simpler. The baseline system used the same network as the SI model, but it was trained independently. During adaptation, we updated the SI model using 5 iterations with a fixed learning rate of $2\times 10^{-4}$ per sample following our previous setup~\cite{lu2016sequence}. We also compared the CE loss \eqref{eq:ce} and the teacher-student loss \eqref{eq:kd} for adaptation (Table \ref{tab:adap}).  When using the CE loss function for both SI models, slightly better results wer obtained when updating the gates only, while updating all the model parameters gave smaller improvements, possibly due to overfitting.  Interestingly, this is not the case for the teacher-student loss, where updating all the model parameters yielded lower WER. These results are also in line with the argument in~\cite{hinton2015distilling} that the soft targets can work as a regularizer and can prevent the student model from overfitting. %To summarise, with less than 0.8 million parameters, the baseline DNN model can obtain 30.8\% WER, while standalone HDNN-$H_{128}L_10$ model with adaptation achieves 28.7\% WER on the {\tt eval} set after sequence training -- 1.6\% absolute better than the corresponding DNN model. With teacher-student training, the WER was further reduced to 28.4\%. 

\subsection{Summary}

We summarize our key results in Table \ref{tab:summary}. Overall, the HDNN acoustic model can slightly outperform the sequence trained baseline using around 5 million model parameters after adapting the gate functions;  using fewer than 2 million model parameters it performed slightly worse. If fewer than 0.8 million parameters are used, then the gap is much larger compared to the DNN baseline. With adaptation and teacher-student training, we can close the gap by around 50\%, with difference in WER falling from roughly 5\% absolute to 2.5\% absolute.

\section{Conclusions}

Highway deep neural networks are structured, depth-gated feedforward neural networks. In this paper, we studied sequence training and adaptation of these networks for acoustic modeling. In particular, we investigated the roles of the parameters in the hidden layers, gate functions and classification layer in the case of sequence training. We show that the gate functions, which only account for a small fraction of the whole parameter set, are able to control the information flow and adjust the behavior of the neural network feature extractors. We demonstrate this in both sequence training and adaptation experiments, in which considerable improvements were achieved by only updating the gate functions. Using these techniques, we obtained comparable or slightly lower WERs with much smaller acoustic models compared to a strong baseline set by a conventional DNN acoustic model with sequence training. Since the number of model parameters is still relatively large compared to the amount of data typically used for speaker adaptation, this adaptation technique may be more applicable to domain adaptation, where the expected amount of adaptation data is larger.

Furthermore, we also investigated teacher-student training for small-footprint acoustic models using HDNNs. We observed that the accuracy of the student acoustic model could be improved under the supervision of a high accuracy teacher model, even without additional unsupervised data. In particular, the student model supervised by an sMBR-based teacher model achieved lower WER compared to the model trained independently using the sMBR-based sequence training approach. Unsupervised speaker adaptation further improved the recognition accuracy by around 5\% relative for a model with fewer then 0.8 million model parameters.  However, we did not obtain improvements either using a hybrid loss function which interpolates the CE and teacher-student loss functions, or using a higher temperature to smooth the pseudo-labels. In the future, we shall evaluate this model in low resource conditions where the amount of training data is much smaller. 

\section{Acknowledgement}

We thank the NVIDIA Corporation for the donation of a Titan X GPU, and the anonymous reviewers for insightful comments and suggestions that helped to improve the quality of this paper.

\bibliographystyle{IEEEbib}
\bibliography{bibtex}

\begin{IEEEbiography}[{\includegraphics[width=1in,height=1.25in,clip,keepaspectratio]{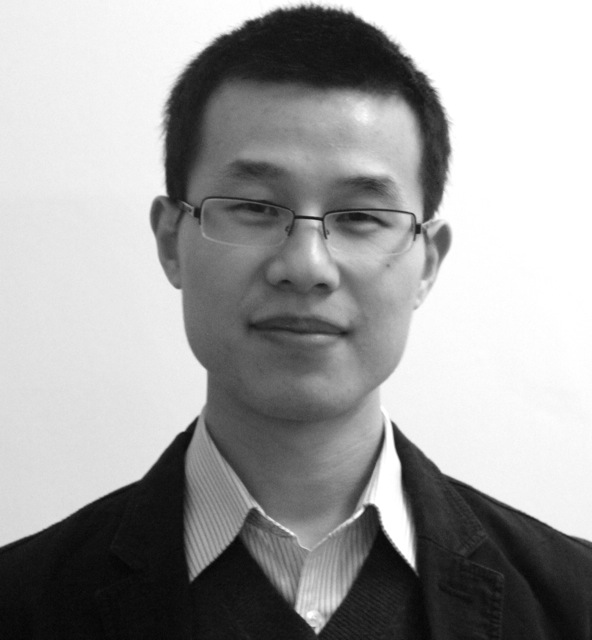}}]{Liang Lu}
a Research Assistant Professor at the Toyota Technological Institute at Chicago. He received his Ph.D. degree from the University of Edinburgh in 2013, where he then worked as a Postdoctoral Research Associate until 2016 before moving to Chicago. He has a broad research interest in the field of speech and language processing. He received the best paper award for his work on the low-resource pronunciation modeling at the 2013 IEEE ASRU workshop. 
\end{IEEEbiography}

\begin{IEEEbiography}[{\includegraphics[width=1in,height=1.25in,clip,keepaspectratio]{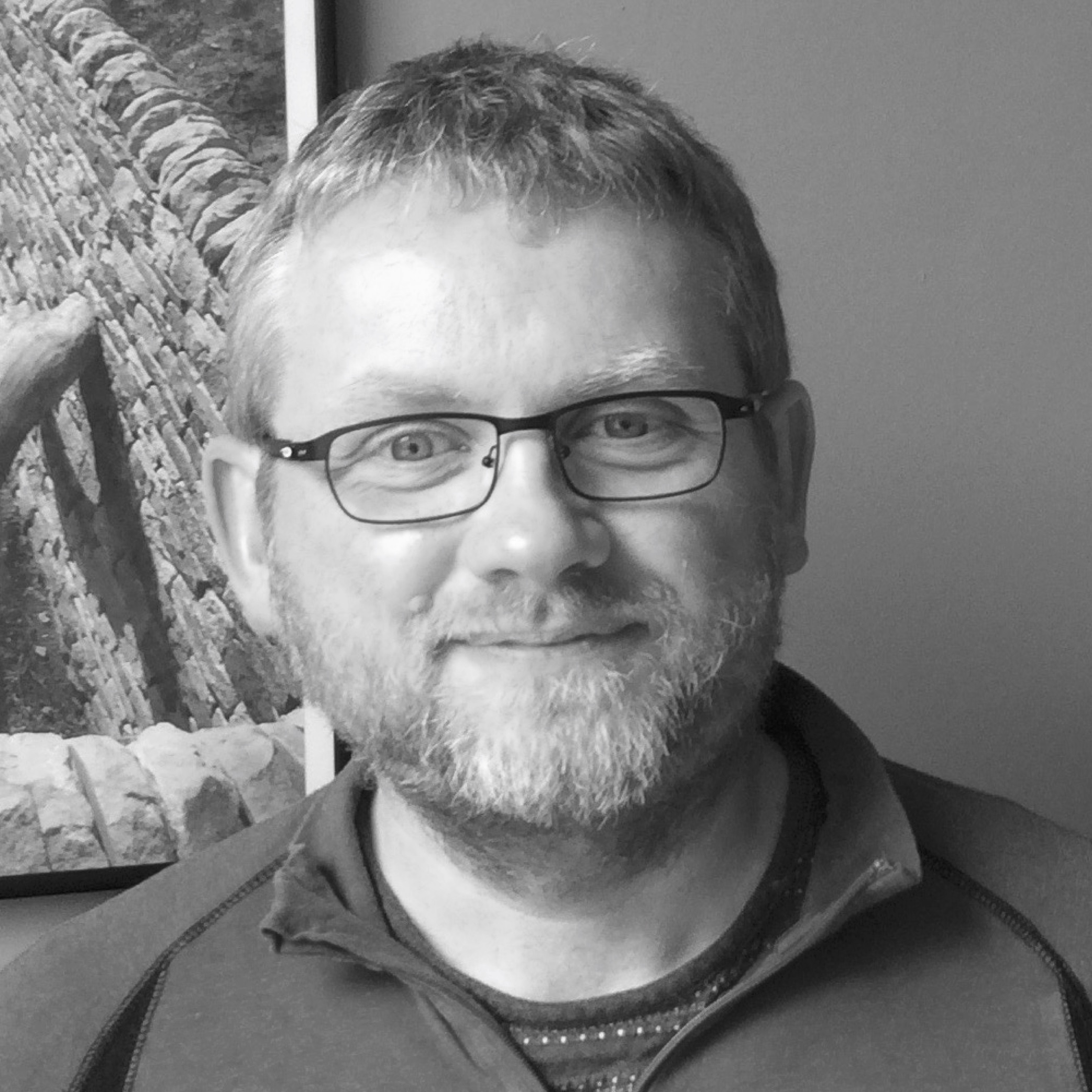}}]{Steve Renals}  (M'91 --- SM'11 --- F'14) 
received the B.Sc. degree from the University of Sheffield, Sheffield, U.K., and the M.Sc. and Ph.D. degrees from the University of Edinburgh. He is Professor of Speech Technology at the University of Edinburgh, having previously had positions at ICSI Berkeley, the University of Cambridge, and the University of Sheffield.  He has research interests in speech and language processing.  He is a fellow of ISCA.
\end{IEEEbiography}

\end{document}